\documentclass[]{imag-ms-template}

\title{Bayesian Uncertainty Quantification for fMRI Functional Connectivity via Simulation-Based Inference}

\author{Simon Carter,$^{1}$ Zeming Kuang,$^{2}$ Lilianne R. Mujica-Parodi,$^{2,3}$ Helmut H. Strey$^{2\ast}$\\
{\small $^{1}$Applied Mathematics and Laufer Center for Physical and Quantitative Biology,}\\
{\small Stony Brook University, Stony Brook NY 11794-5281, USA}\\
{\small $^{2}$Biomedical Engineering Department and Laufer Center for Physical and Quantitative Biology,}\\
{\small Stony Brook University, Stony Brook NY 11794-5281, USA}\\
{\small $^{3}$Santa Fe Institute, Santa Fe, NM, USA}\\
{\small $^\ast$Correspondence: helmut.strey@stonybrook.edu}
}

\begin{document}

\maketitle


\begin{abstract}
Optimizing fMRI scan duration and spatial resolution is critical for experimental 
design, yet traditional correlation-based approaches cannot quantify uncertainty or 
disentangle scanner measurement noise from true neural variability across subjects. 
Without principled uncertainty bounds, researchers cannot know whether a given protocol 
is long enough to reliably estimate connectivity, or whether observed between-subject 
differences reflect genuine biological variation or noise. We present a Bayesian 
framework modeling BOLD dynamics as coupled Ornstein-Uhlenbeck processes, using 
Sequential Neural Posterior Estimation to obtain connectivity posteriors while properly 
accounting for frequency-independent measurement noise that contaminates the entire BOLD 
spectrum. Applied to $N = 28$ healthy controls (55 scans) at 7T using a functional 
network atlas (65 DMN regions), this framework provides a principled tool to optimize 
acquisition parameters by quantifying uncertainty across its sources: scanner noise, 
subject variability, and acquisition length. Spatial analysis identifies a mean of 46 
voxels per ROI, roughly half of typical region sizes, as sufficient to achieve 90\% of 
asymptotic precision. At the single-subject level, 7T reaches its within-session 
precision plateau in approximately 7 minutes compared with 10 minutes for 3T, a 
40\% reduction in required scan time, providing the first direct, model-based 
quantification of the scan-time advantage conferred by higher field strength. At the 
population level, 7T maintains its advantage: 3T requires roughly 37 times more 
per-subject scan time than 7T for the pooled curves to converge, confirming a 
consistent advantage of higher field strength at every timescale. Together these 
findings provide concrete, scanner-specific guidance for protocol optimization, with 
direct implications for reducing acquisition costs and improving the reliability of 
connectivity-based clinical biomarkers. We provide code enabling researchers to derive 
these bounds from their own data.
\end{abstract}

\section{Introduction}
Resting-state functional MRI (rs-fMRI) has established that the brain organizes into 
reproducible large-scale networks whose intrinsic connectivity correlates with behavior, 
cognition, and disease \cite{Biswal1995}. A substantial body of work has characterized 
how reliably these connectivity estimates can be measured. Test-retest studies show that 
edgewise correlations stabilize after approximately 5--15 minutes of acquisition, with 
the exact plateau depending on field strength, TR, spatial resolution, and the network 
under study \cite{Birn2013, van_dijk_intrinsic_2010, Noble2019}. Pannunzi et al.\ 
showed that finite-sample variance decays as a power law in timepoints before falling 
below within-subject variance \cite{Pannunzi2017}, and Finn et al.\ demonstrated that 
individual subjects can be identified from their connectivity patterns with relatively 
short scans \cite{Finn2015}. More recently, Ma et al.\ showed that advanced connectivity techniques such as dynamic causal modeling can achieve acceptable reliability with scanning durations as short as 10.8 minutes and sample sizes as few as 40 subjects~\cite{Ma2024}, further underscoring the practical importance of understanding the relationship between acquisition parameters and estimation precision.
Together these findings have produced empirical benchmarks 
that now shape standard protocols, including the Human Connectome Project's 15-minute 
standard \cite{smith_resting-state_2013}.

These benchmarks, however, share a common limitation: they are derived empirically, 
through repeated scanning sessions or split-half designs, rather than from first 
principles applied to a single acquisition \cite{Noble2019, Noble2021}. Deriving 
reliability benchmarks requires large test-retest datasets that are themselves expensive 
to collect, somewhat undermining the goal of reducing acquisition costs given that a 
single hour of scanner time can exceed \$600 at typical academic rates 
\cite{van_dijk_intrinsic_2010}. More fundamentally, neither Pearson correlation nor 
ICC-based reliability measures can separate scanner measurement noise from true 
between-subject neural variability \cite{Liu2016}. When two subjects show different 
connectivity estimates, current methods cannot determine whether the difference reflects 
genuine biological heterogeneity or differential noise contamination \cite{Murphy2013}.

This inability to attribute observed differences to their source has direct clinical 
consequences. Without principled uncertainty bounds, researchers cannot know whether a 
given protocol is long enough to reliably estimate connectivity, or whether apparent 
between-subject differences reflect genuine biology or noise. Conservative scanning 
protocols are therefore typically adopted not because they are known to be necessary, 
but because the cost of being wrong cannot be quantified. This is a fundamental barrier 
to using connectivity as a single-subject clinical biomarker \cite{Finn2015}, where 
group averaging is unavailable and decisions must rest on individual measurements.

Here we present a Bayesian framework that addresses these limitations directly, by 
assuming an underlying generative model and performing Bayesian parameter estimation 
\cite{friston_classical_2002, friston_variational_2007, strey_estimation_2019, 
carter_parameter_2024}. Using coupled Ornstein-Uhlenbeck processes to capture BOLD 
signal dynamics while accounting for frequency-independent measurement noise, we derive 
principled uncertainty bounds for connectivity estimates from a single acquisition 
session without requiring repeated scanning. The framework quantifies uncertainty 
arising from each of three sources in turn: scanner measurement noise, acquisition 
length, and between-subject variability. By comparing Bipolar (7T) and HCP-YA (3T) 
datasets we provide a direct, model-based quantification of the scan-time advantage 
of higher field strength: 7T requires less scan time than 3T at every level of 
comparison, with the advantage growing stronger at the population level where 
higher per-subject precision compounds across subjects. We validate 
our approach using regions from the default mode network \cite{raichle_default_2001}, 
a well-characterized system whose intrinsic activity and clinical relevance have made 
it a benchmark for connectivity methods.

\section{Coupled Ornstein-Uhlenbeck Model}
We model the underlying neural signal dynamics as a system of coupled Ornstein-Uhlenbeck 
processes. Since physical coupling naturally produces positive correlations (like two 
masses connected by a spring), we introduce a scaling factor to allow for both positive 
and negative correlations in the observed signals:
\begin{align}
    dx_1(t) &= [-\theta_1 x_1(t) + \nu (x_2^*(t) - x_1(t))]dt + \sigma_1 dW_1(t), \\
    dx_2^*(t) &= [-\theta_2 x_2^*(t) + \nu (x_1(t) - x_2^*(t))]dt + \sigma_2 dW_2(t), \\
    x_2(t) &= \gamma \cdot x_2^*(t)
\end{align}
where $W_1(t)$ and $W_2(t)$ are independent standard Brownian motions, $\theta_1, 
\theta_2 > 0$ are mean-reversion parameters, $\nu \geq 0$ represents the coupling 
strength, and $\sigma_1, \sigma_2 > 0$ are volatility parameters. Here, $x_2^*(t)$ 
represents the intrinsic dynamics of region 2, while the observed signal $x_2(t)$ 
includes a scaling factor $\gamma \in \mathbb{R}$ that can be negative, allowing the 
system to exhibit anticorrelated behavior despite positive coupling strength $\nu$. This 
parameterization has biological interpretability: $\nu$ represents synaptic strength 
(always positive), while $\gamma$ determines whether the connection is excitatory 
($\gamma > 0$) or inhibitory ($\gamma < 0$). We place a normal prior $\gamma \sim 
\mathcal{N}(0, \sigma_\gamma^2)$ to explore both correlation regimes equally.

The observed measurements $y_1(t)$ and $y_2(t)$ include additional independent 
Gaussian measurement noise:
\begin{align}
    y_i(t) = x_i(t) + \epsilon_i(t), \quad \epsilon_i(t) \sim \mathcal{N}(0, \phi_i^2), 
    \quad i = 1,2
\end{align}
This parameterization yields eight model parameters: $\boldsymbol{\theta} = (\theta_1, 
\theta_2, \nu, \sigma_1, \sigma_2, \phi_1, \phi_2, \gamma)$.

The coupling between processes generates a complex correlation structure that depends 
non-linearly on all eight parameters. We define the noiseless correlation as the 
zero-lag cross-correlation between the true signals $x_1(t)$ and $x_2(t)$, excluding 
the contribution of measurement noise $\epsilon_i(t)$. To recover this quantity from 
the full parameter posterior, we train a separate decoder network to approximate the 
mapping from parameter space to noiseless correlation. Parameters are log-transformed 
prior to training, $\tilde{\theta}_{1:7} = \log(\theta_{1:7})$, with $\tilde{\theta}_8 
= \theta_8$, and the regression target is the noiseless correlation after a clipped 
Fisher $z$-transform,
\begin{equation}
    z = \operatorname{arctanh}(\operatorname{clip}(r,\, -1+\epsilon,\, 1-\epsilon)), 
    \quad \epsilon = 0.01.
\end{equation}
The decoder is a fully connected MLP with architecture $8 \to 32 \to 16 \to 1$, ReLU 
activations in hidden layers, and a linear output. It was trained using the Adam 
optimizer (learning rate $0.01$) with mean squared error loss for 50,000 epochs, and 
exported as TorchScript for downstream use.

To make parameter estimation computationally tractable, we reduce the dimensionality 
of the observed data by extracting seven summary statistics that capture the essential 
correlation structure:
\begin{itemize}
    \item Autocorrelation functions: $R_{11}(\tau_k)$ for $k = 1,2,3$ where $\tau_k 
    \in \{1, 5, 10\}$ time steps
    \item Autocorrelation functions: $R_{22}(\tau_k)$ for $k = 1,2,3$ 
    \item Zero-lag cross-correlation: $R_{12}(0)$
\end{itemize}
This choice of summary statistics is motivated by their ability to capture both the 
individual process dynamics (through autocorrelation) and the coupling strength 
(through cross-correlation), while maintaining computational efficiency.

All simulated and observed signals undergo identical preprocessing, including 
fifth-order Butterworth bandpass filtering (0.01--0.1 Hz), consistent with standard 
fMRI preprocessing pipelines implemented in nilearn \cite{abraham_machine_2014}.

\section{SBI Methodology}
Traditional band-pass filtering is commonly used to address noise in fMRI data, but 
because noise spans the entire frequency spectrum, residual noise inevitably remains 
in the filtered signal. While averaging across trials can reduce this remaining noise, 
such an approach becomes problematic when examining single-subject metrics needed for 
clinical diagnosis using fMRI. Simulation-based inference (SBI) offers an alternative: 
by explicitly modeling the underlying generative process and noise characteristics, we 
can more effectively separate signal from noise and obtain principled uncertainty 
estimates for individual subjects.

We employ Amortized Neural Posterior Estimation (NPE) to fit the parameters of the 
coupled Ornstein-Uhlenbeck model described previously. Traditional Bayesian inference 
requires evaluating a likelihood function, which for complex dynamical systems like 
ours is often intractable or computationally prohibitive \cite{cranmer_frontier_2020}. 
Likelihood-free methods such as Approximate Bayesian Computation (ABC) circumvent this 
limitation by comparing simulated data to observations, but require running new 
simulations for each inference query, a computational cost that becomes prohibitive 
when analyzing thousands of region pairs across multiple subjects. NPE addresses this 
through amortization: a neural density estimator is trained once on a large set of 
simulations, learning to map directly from summary statistics of observed data to the 
posterior distribution over model parameters \cite{greenberg_automatic_2019, 
Azukas2026}. Once trained, inference on new observations requires only a single forward 
pass through the network, enabling the rapid pairwise analysis necessary for 
whole-brain connectivity studies.

To train the neural density estimator, we drew 500,000 simulations from the joint 
prior, each producing a time series of 584 seconds sampled at 1.25 Hz. The posterior 
was conditioned on the seven summary statistics described above, with the decoder 
trained separately to predict the noiseless correlation as generated by the coupled OU 
model. The posterior was estimated using NPE as implemented in the \texttt{sbi} 
package \cite{papamakarios_normalizing_2021}, with default training hyperparameters 
and a normalizing flow density estimator. The trained posterior object was serialized 
to disk, enabling rapid conditional sampling for any new observation without retraining.

\section{Data and Preprocessing}

\subsection{Bipolar Dataset (7T)}
Resting-state fMRI data were acquired from $N = 28$ healthy controls (55 scans) at a 7 Tesla MRI 
scanner (MAGNETOM Terra, Siemens Healthcare, Erlangen, Germany) equipped with a custom-build 64-channel headcoil at the Massachusetts General Hospital Athinoula A. Martinos Center for 
Biomedical Imaging, as part of the PAgB study~\cite{mujica-parodi_diet_2020}. Informed consent was obtained from all participants for being included in this study.  BOLD 
images were acquired using the following protocol: simultaneous multi-slice (SMS) slice 
acceleration factor = 5, $R = 2$ acceleration in the primary phase encoding direction 
(62 reference lines), TR = 802 ms, TE = 20 ms, flip angle = 33°, voxel size = 
$1.75 \times 1.75 \times 1.75$ mm, 85 slices, and 730 measurements ($\approx$10 
minutes). Subject-level T1-weighted and fMRI images were preprocessed using fMRIPrep 
(version 20.2.3)~\cite{Esteban2019}. Confound removal, bandpass filtering 
(0.01--0.1 Hz, 5th-order Butterworth), detrending, and standardization were performed 
using Nilearn~\cite{abraham_machine_2014}. Framewise displacement (FD) was computed 
for each scan; scans with mean FD $> 0.5$ mm were excluded from analysis. White 
matter, CSF, and motion parameters (three translations and three rotations) were 
included as confound regressors. The default mode network was parcellated using the 
Seitzman atlas~\cite{seitzman_set_2020}, yielding 65 regions of interest.

\subsection{HCP Young Adult Dataset (3T)}
Data from the Human Connectome Project Young Adult (HCP-YA) 
dataset~\cite{van_essen_wu-minn_2013} were used for cross-field-strength comparison. 
We used the standard HCP minimally preprocessed resting-state fMRI data, as described 
in Glasser et al.~\cite{glasser_minimal_2013}. The same Seitzman atlas parcellation 
and bandpass filtering (0.01--0.1 Hz) were applied to the HCP-YA timeseries to ensure 
comparability with the 7T dataset.

\section{Validation on Synthetic Data}
\begin{figure}[htbp]
\begin{center}
\includegraphics[width=0.8\textwidth]{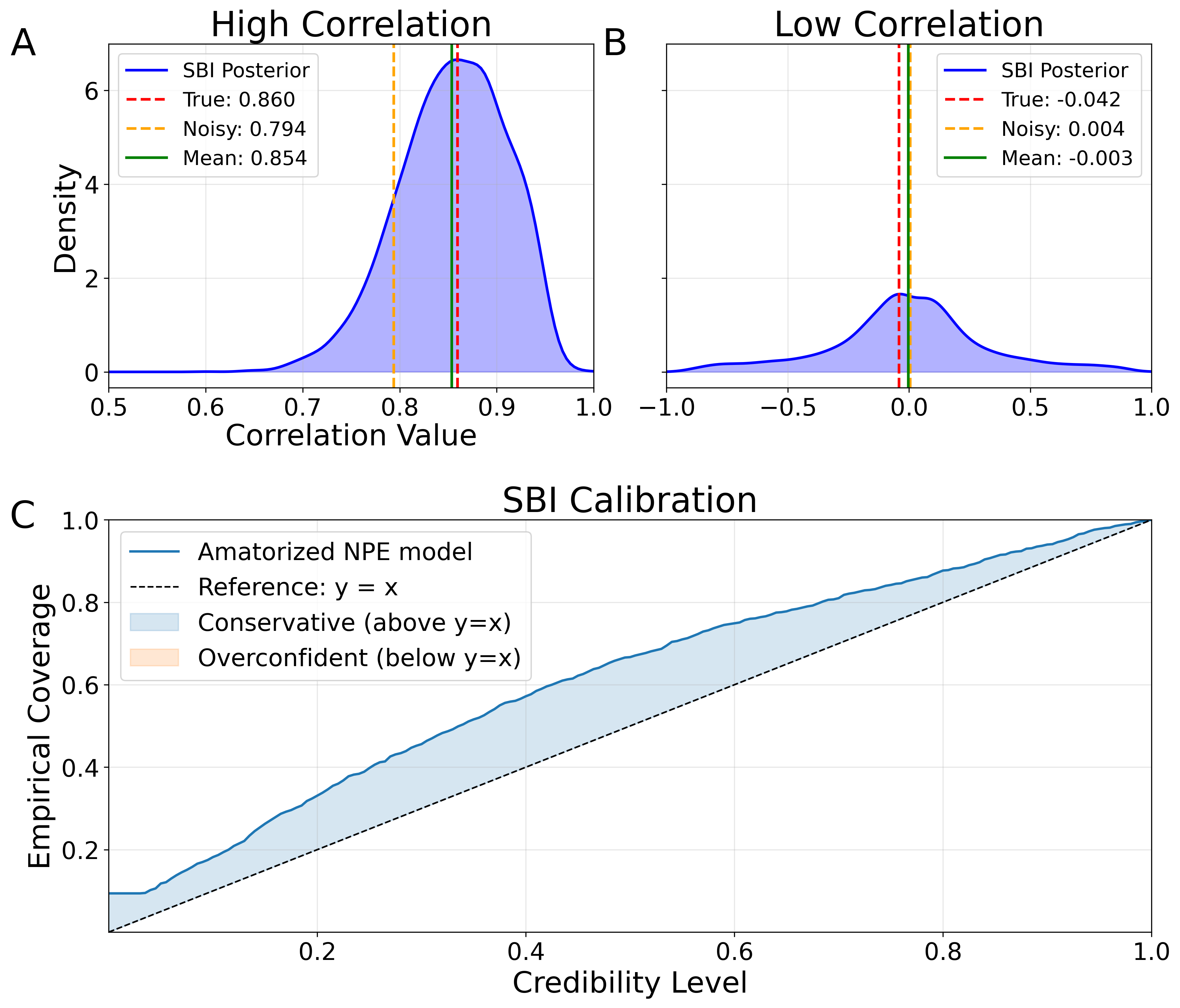}
\caption{\textbf{A.} Example posterior distribution from 
the toy model for a case with high true correlation ($\rho > 0.5$). The posterior 
distribution (blue) is centered close to the true correlation value (red dashed line), 
with the noisy Pearson correlation coefficient (orange dashed line) and posterior mean 
(green solid line) also indicated. \textbf{B.} The equivalent case for low true 
correlation ($|\rho| < 0.1$), showing the posterior distribution (blue) again centered 
near the true value (red dashed line). \textbf{C.} A calibration plot showing empirical 
coverage versus credibility level. The curve lying above the diagonal reference line 
($y = x$) indicates conservative posterior distributions, with actual coverage exceeding 
nominal credibility levels.}
\label{fig:toy_validation}
\end{center}
\end{figure}

To validate our approach, we generated synthetic data from coupled Ornstein-Uhlenbeck (OU) processes, which have been established as appropriate models for resting-state neuroimaging signals~\cite{Gilson2020}. The OU process captures key statistical properties of band-limited neural recordings, including stationary autocorrelation structure and coupling-induced correlations~\cite{Deco2013}. We simulated 1,000 band-pass filtered coupled OU processes and compared our SBI model's performance against traditional Pearson correlation.

The SBI model was trained on 100,000 simulations with uniform priors over biologically plausible parameter ranges, except for the coupling parameter, which was sampled from an exponential distribution to yield approximately uniform correlation values in the filtered signals.

While our method achieved modest improvements in point estimation accuracy, 1.5\% overall (95\% CI: 0.94--2.02\%) and 3.42\% for $|\rho| \geq 0.2$ (95\% CI: 2.79--4.05\%), the primary advantage lies in providing full posterior distributions that enable principled uncertainty quantification (Figure~\ref{fig:toy_validation}A--B). Traditional correlation methods offer no such capability.

To assess the reliability of these uncertainty estimates, we evaluated posterior calibration following established procedures~\cite{talts_validating_2020, hermans_trust_2022}. For a well-calibrated model, the empirical coverage, the proportion of ground truth values falling within highest posterior density (HPD) intervals, should match the nominal credibility level. Our calibration analysis (Figure~\ref{fig:toy_validation}C) reveals that empirical coverage consistently exceeds the credibility level, with an integrated calibration error of 0.1125 indicating conservative uncertainty estimates. This conservative bias means our posterior credible intervals are wider than strictly necessary, providing reliable uncertainty bounds that will not understate estimation uncertainty when applied to empirical neural data.

\section{Validation on Resting-State Data}
We next validated that our method effectively reduces noise in real neuronal data. We analyzed 7 Tesla resting-state fMRI data, examining connectivity within the default mode network (DMN), a network known to exhibit robust intrinsic connectivity during rest \cite{raichle_default_2001}. The DMN was parcellated using the Seitzman atlas~\cite{seitzman_set_2020}, yielding 65 regions of interest. We computed pairwise correlations between all DMN regions from 10-minute resting-state time series. The posterior model was trained specifically for this time series duration and applied to estimate denoised correlations, which were then compared to traditional Pearson correlation coefficients.

If our method successfully removes noise, we should observe two key outcomes. First, DMN correlations should increase relative to Pearson correlation, as uncorrelated measurement noise exclusively attenuates connectivity estimates. Second, the uncertainty estimates provided by our method should enable researchers to push experimental boundaries while maintaining quantifiable data quality. Specifically, uncertainty metrics allow researchers to determine how short they can make their scanning protocols or how fine-grained their spatial parcellations can be while still achieving acceptable measurement precision. This capability addresses a critical limitation in current practice, where conservative scanning parameters are often adopted due to the absence of subject-specific quality metrics, potentially leading to unnecessarily long scans or overly coarse spatial resolutions.

\subsection{Voxel by Voxel Validation}
To further demonstrate our method's uncertainty quantification capabilities, we examined 
how the posterior distribution behaves with reduced spatial averaging, using a single 
representative subject from the bipolar dataset. In conventional Pearson correlation, 
individual voxels suffer from noise that attenuates observed correlations, and researchers 
lack a principled way to assess the reliability of their estimates, particularly at the 
single-subject level, where group averaging cannot compensate for measurement uncertainty.

To perform inference, we trained a neural posterior estimator using simulations generated from one million parameter sets, each producing a 584-second time series sampled at 1.25 Hz. We employed the same prior distributions as the toy model, resulting in uniformly distributed correlation priors. To ensure unconstrained sampling during inference, correlations were sampled in Fisher $z$-transformed space and then back-transformed via the inverse hyperbolic tangent ($\tanh$) to the correlation space, constraining values to the interval $[-1, 1]$.

We applied this trained estimator to empirical data from the Default Mode Network, examining connectivity estimates across all region pairs (2080 pairs total). For each pair, we calculated the number of voxels required to reach 90\% of the final correlation estimate and assessed reliability using the ratio of posterior mean to posterior standard deviation, where values greater than one indicate that the estimated correlation exceeds its uncertainty.

\begin{figure}[htbp]
\begin{center}
\includegraphics[width=0.8\textwidth]{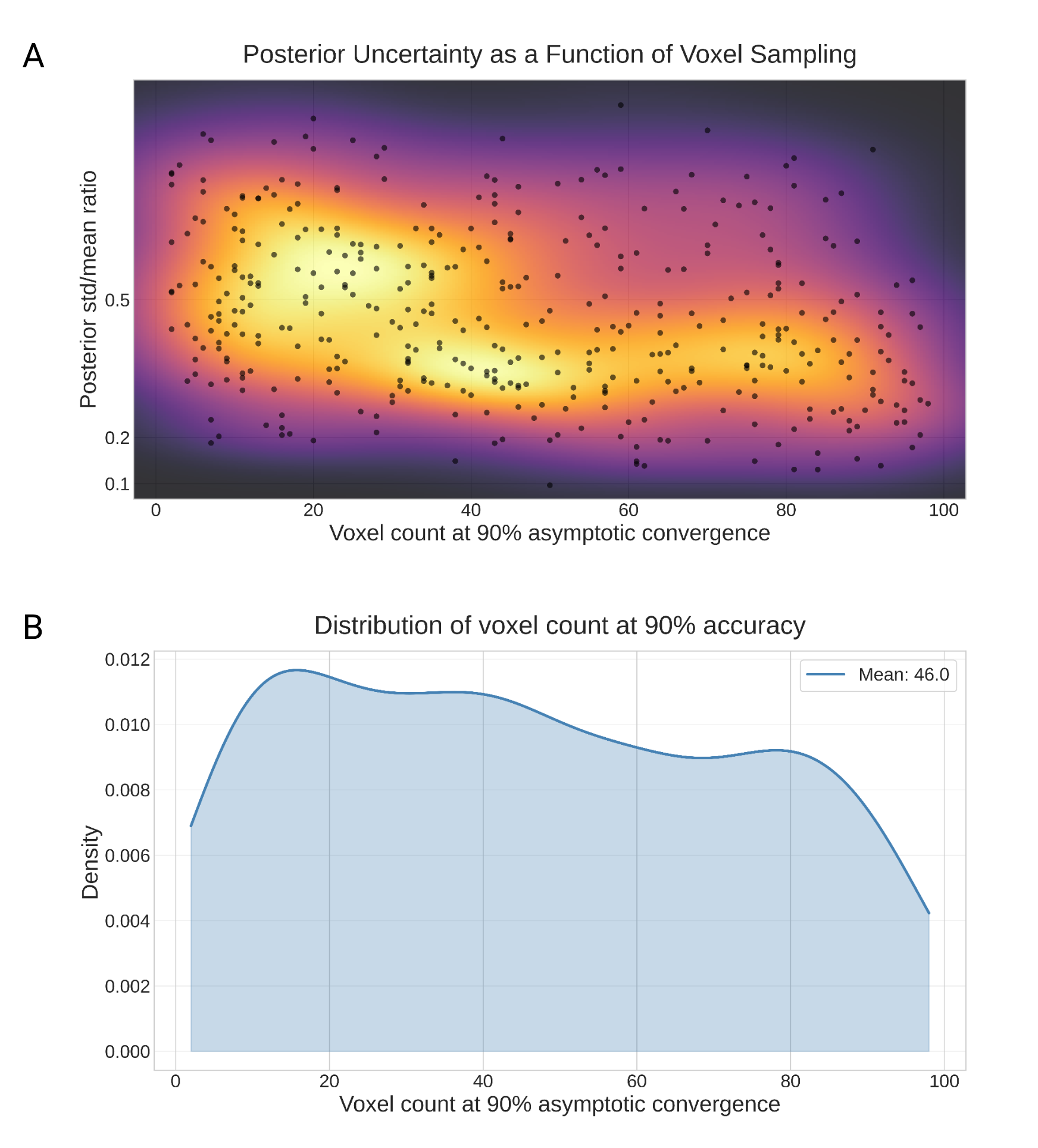}
\caption{\textbf{A.} Each point represents a pair of 
regions within the Default Mode Network. Of 2,080 total candidate region pairs, 1,302 
were removed for failing a voxel homogeneity test ($R^2 < 0.5$), which assesses whether 
the correlation estimate behaves asymptotically as voxels are added. The remaining 411 
pairs are those where the ratio of the posterior standard deviation to the posterior mean 
was less than 1, indicating sufficient confidence in the correlation estimate. The x-axis 
shows the number of voxels required to reach 90\% of the final correlation estimate 
(obtained by averaging all voxels), and the y-axis shows this ratio, where lower values 
indicate greater confidence that a true correlation exists. Notably, many region pairs 
achieve reliable estimates with half or fewer of the total voxels. \textbf{B.} Density 
distribution of the number of voxels at which each region pair reached 90\% of its 
asymptotic correlation estimate, across the 411 pairs shown in \textbf{A}. The 
approximately uniform distribution, with a mean of 46 voxels, indicates that the minimum 
number of voxels required for stable estimates is highly region-specific rather than 
governed by a single global threshold.}
\label{fig:voxel_reduction}
\end{center}
\end{figure}

We first applied a voxel homogeneity test to identify pairs where signal is distributed evenly across voxels. For each region pair, we fit a saturating exponential function to the correlation estimates as voxels were incrementally added; pairs that conformed well to this asymptotic model were considered homogeneous, indicating that each voxel contributes similarly to the overall signal. Pairs that failed this test (1302, 63\%) were excluded from this particular analysis. Importantly, this exclusion does not indicate that our Bayesian method is inapplicable to these pairs, uncertainty quantification via the posterior standard deviation remains valid. Rather, the voxel-reduction analysis assumes that each voxel within an ROI contributes a common signal, and this assumption is violated when averaging is dominated by a small subset of voxels or when voxels represent heterogeneous signals. The high exclusion rate likely reflects genuine biological heterogeneity within anatomically defined regions rather than a limitation of the method itself. This left 746 pairs for subsequent analysis (Figure~\ref{fig:voxel_reduction}A).

Among these pairs, we observed that many achieved stable estimates with half or fewer of the total voxels, confirming that our uncertainty quantification approach reduces the voxel burden required for reliable connectivity measurement. To illustrate how this framework can inform thresholding decisions, we applied a cutoff on the ratio of posterior standard deviation to posterior mean of 1, removing all pairs for which the estimated correlation does not exceed its own uncertainty. This left 411 pairs (Figure~\ref{fig:voxel_reduction}A). Figure~\ref{fig:voxel_reduction}B shows the distribution of voxel counts at which these pairs reached 90\% of their asymptotic correlation estimate; the distribution is approximately uniform across all 411 pairs, indicating that the minimum number of voxels required for stable estimates is highly region-specific rather than governed by a single global threshold. Together, these results, derived from a single subject, demonstrate that our method provides meaningful uncertainty quantification for connectivity estimates at the single-subject level. By quantifying uncertainty, researchers gain two practical capabilities: first, they can calibrate the spatial specificity of their ROIs by identifying the minimum number of voxels needed for reliable connectivity measurement while preserving anatomical precision; second, the posterior distribution provides a principled, non-arbitrary basis for determining which connections have sufficient evidence to be considered reliable, without relying on group statistics or arbitrary thresholds.

\subsection{Optimizing Scan Duration for Dynamic Connectivity Studies}

\begin{figure}[htbp]
\begin{center}
\includegraphics[width=0.8\textwidth]{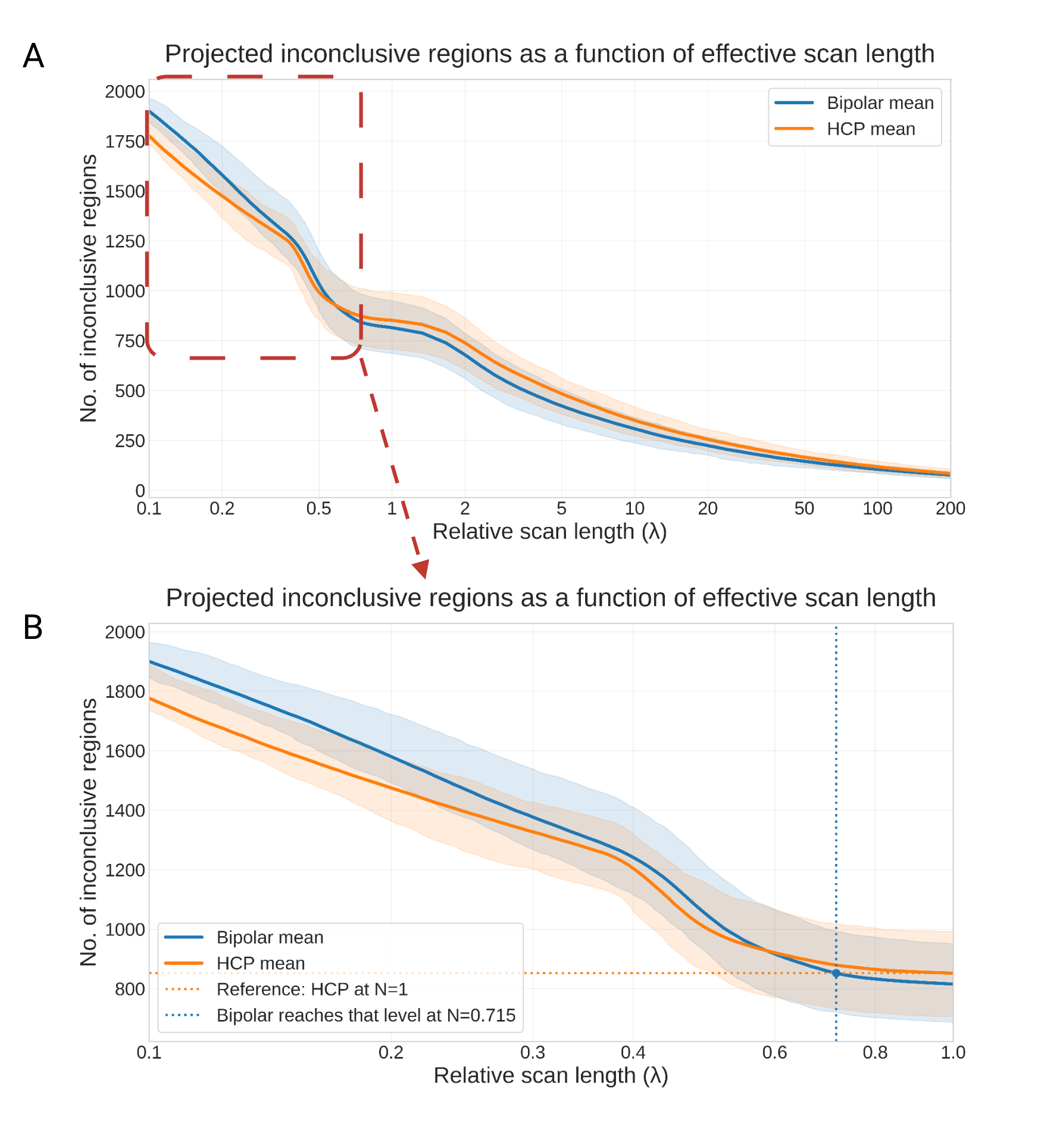}
latex\caption{\textbf{A.} Average number of inconclusive region
pairs as a function of relative scan length $\lambda$ for the Bipolar (7T) and HCP-YA
(3T) datasets, extended to the extrapolated global regime. Here $\lambda = 1$ corresponds
to the full 730-timepoint Bipolar acquisition ($\approx$10 minutes). Bipolar (7T)
matches the conclusiveness that HCP-YA (3T) attains at
$\lambda_{\mathrm{eq}}^{\mathrm{glob}} = 200$ with only $\lambda \approx 162.38$,
confirming that higher field strength requires less scan time to achieve equivalent
precision. Both global thresholds fall far outside any practical session, shown here to
establish the theoretical precision ceiling of each field strength. Regions are defined
by the Seitzman atlas (2,080 pairs).
\textbf{B.} Local within-session view, restricted to $\lambda \in [0.1, 1]$. Bipolar (7T)
matches the conclusiveness of the full 10-minute HCP-YA (3T) acquisition at
$\lambda_{\mathrm{eq}}^{\mathrm{loc}} = 0.715$ (approximately 7 minutes). A 7T scan of
only 7 minutes therefore achieves the same within-session precision as a full 10-minute
3T scan.}
\label{fig:scan_duration}
\end{center}
\end{figure}

A fundamental challenge in resting-state fMRI study design is determining optimal scan duration \cite{duda_reliability_2023, abdul_wahab_effects_2022}. While longer scans generally improve the reliability of connectivity estimates \cite{van_dijk_intrinsic_2010}, they also increase participant burden, motion artifacts, and scanner costs. Traditional approaches to optimizing scan parameters require large test-retest datasets, limiting their practical utility.

Our approach enables scan duration optimization from a single subject. By exploiting the Bayesian uncertainty, we can characterize how estimate precision changes as a function of acquisition length. We introduce $\lambda$ to denote relative scan length, where $\lambda = 1$ corresponds to the full 730-timepoint Bipolar acquisition ($\approx$10 minutes) and $\lambda < 1$ corresponds to a proportionally shorter scan. We exploit the fact that the posterior standard deviation should decrease approximately at a rate of $1/\sqrt{n}$, where $n$ is the number of observations. For each region pair, let $\{\theta_s\}_{s=1}^{S}$ denote the current posterior samples with empirical mean $\hat{\mu}$. To approximate the posterior under a hypothetical dataset $\lambda$ times as long, we apply the rescaling:
\begin{equation}
    \theta_s^{(\lambda)} = \hat{\mu} + \frac{\theta_s - \hat{\mu}}{\sqrt{\lambda}}
\end{equation}
This is motivated by the Bernstein--von Mises theorem, under which the posterior concentrates around the true parameter at rate $1/\sqrt{n}$ under regularity conditions, converging to a Gaussian with variance proportional to $1/n$. Consequently, if the current posterior reflects $n$ observations, the posterior based on $\lambda n$ observations would have standard deviation reduced by a factor of $\sqrt{\lambda}$. The rescaling above applies this contraction while preserving the empirical shape of the distribution; no parametric form is assumed. This approximation is exact when the posterior is Gaussian and serves as a first-order approximation otherwise. Note that the Bernstein--von Mises theorem does not generally hold in nonparametric settings; in such cases the rescaling should be interpreted as a heuristic motivated by the parametric asymptotic regime.

For each dataset we identify two thresholds, both defined as the scan length at which the Bipolar (7T) and HCP-YA (3T) conclusiveness curves intersect. The \emph{local} threshold $\lambda_{\mathrm{eq}}^{\mathrm{loc}}$ is the intersection point within the observable session range ($\lambda \leq 1$), identifying the within-session scan length at which the two datasets achieve equal conclusiveness. The \emph{global} threshold $\lambda_{\mathrm{eq}}^{\mathrm{glob}}$ is the intersection point in the extrapolated large-$\lambda$ regime, identifying the scan length at which the two datasets' projected curves converge when extended far beyond a single session. Comparing these intersection points directly quantifies how much less scan time 7T requires to match 3T precision at each timescale.

The resulting figures (Figure~\ref{fig:scan_duration}) demonstrate the 7T advantage quantitatively and at both levels. At the local level, Bipolar (7T) reaches its within-session plateau at $\lambda_{\mathrm{eq}}^{\mathrm{loc}} = 0.715$ (approximately 7 minutes), while HCP-YA (3T) does not plateau until $\lambda_{\mathrm{eq}}^{\mathrm{loc}} = 1.0$ (its full 10-minute acquisition). A 7T scan of only 7 minutes therefore achieves the same within-session precision as a full 10-minute 3T scan, a reduction of approximately 3 minutes or 28.5\% in required scan time. At the global level, the two projected curves intersect at $\lambda_{\mathrm{eq}}^{\mathrm{glob}} = 162.38$ for Bipolar and $\lambda_{\mathrm{eq}}^{\mathrm{glob}} = 200$ for HCP-YA. Neither value is a practical scanning target, but their comparison establishes that 7T reaches the same far-extrapolated precision level with 18.8\% less data than 3T, and that 7T's precision in this regime is intrinsically higher --- a consequence of the greater BOLD signal amplitude and temporal SNR at higher field strength \cite{triantafyllou2005, ugurbil2012, demartino2018} that cannot be recovered by extending a 3T acquisition.

\subsection{Across Population Effects}

\begin{figure}[htbp]
\begin{center}
\includegraphics[width=0.8\textwidth]{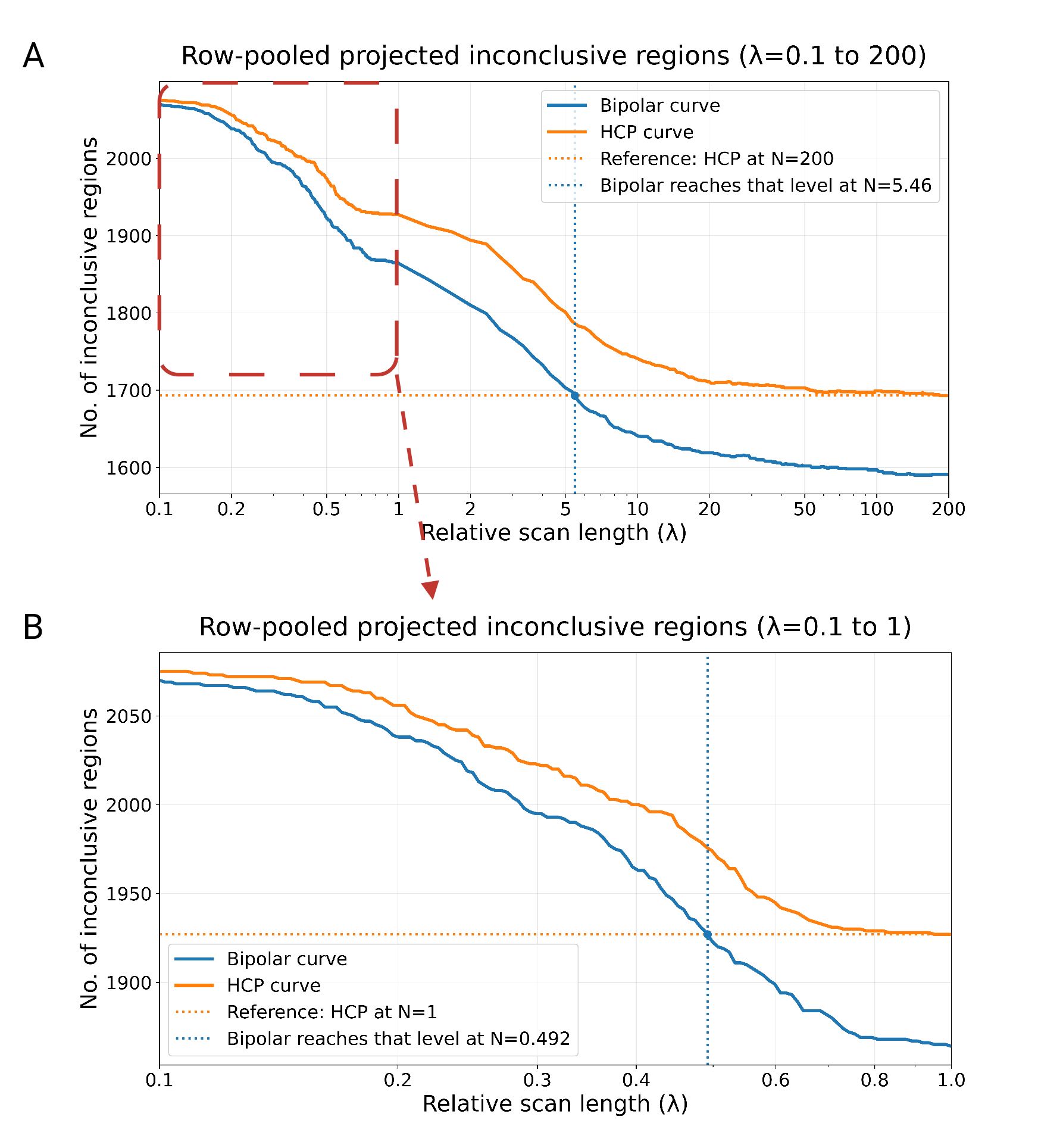}
\caption{\textbf{A.} Proportion of inconclusive region
pairs as a function of number of subjects $N$ for the Bipolar (7T) and HCP-YA (3T)
resting-state datasets. Individual posteriors are rescaled under the $1/\sqrt{n}$
approximation before concatenation to form the population-level distribution. The
Bipolar (7T) cohort consistently yields fewer inconclusive region pairs than HCP-YA
(3T) at equivalent $N$, confirming that the scan-time advantage of 7T extends to and
is amplified at the population level. \textbf{B.} Per-subject scan-length view showing
relative acquisition length $\lambda \in [0.1, 1]$ for the pooled distributions.
Bipolar (7T) matches the conclusiveness of the full HCP-YA (3T) acquisition at
$\lambda_{\mathrm{eq}}^{\mathrm{loc}} \approx 0.492$, meaning 7T requires only about
half the per-subject scan time of 3T to achieve equivalent population-level precision
within session. In the extrapolated global regime, the advantage widens further:
Bipolar (7T) matches the conclusiveness that HCP-YA (3T) attains at
$\lambda_{\mathrm{eq}}^{\mathrm{glob}} = 200$ with only $\lambda \approx 5.46$,
meaning pooled 3T requires roughly 37 times more per-subject scan time than pooled 7T
to reach the same precision.}
\label{fig:pooled_posteriors}
\end{center}
\end{figure}

To assess connectivity at the population level, we adopt a hierarchical Bayesian perspective in which each subject's true connectivity parameter $\theta_i$ is assumed to be drawn from an unknown population distribution $P(\theta)$, with subjects treated as exchangeable \cite{gelman2013bda}. Rather than placing an explicit parametric prior on $P(\theta)$, we estimate it nonparametrically by concatenating posterior samples across subjects \cite{robbins1956}. Under exchangeability, the order of subjects is uninformative, and each subject contributes equally to the pooled distribution. The resulting concatenated samples therefore serve as a nonparametric empirical Bayes estimate of the population distribution $P(\theta)$ \cite{friston2016}.
 
An important consequence of pooling is that concatenating posteriors across subjects introduces between-subject variance into the population-level distribution. This variance reflects genuine heterogeneity across individuals and causes the pooled distribution to be wider than any single subject's posterior. However, because 7T produces individual posteriors with higher precision, the pooled 7T distribution remains sharper than its 3T counterpart even after this variance is introduced \cite{triantafyllou2005, ugurbil2014, hale2010}. As a result, pooled Bipolar (7T) consistently yields \emph{fewer} inconclusive region pairs than pooled HCP-YA (3T) at equivalent group size, and the 7T scan-time advantage is preserved and amplified at the population level.

To assess how scan duration affects population-level connectivity estimates, we apply the same $1/\sqrt{n}$ rescaling of the posterior standard deviation introduced in the previous 
section, but now to the pooled distribution formed by concatenating individual posteriors 
across subjects. Critically, this rescaling is applied to each individual subject's 
posterior prior to concatenation, rather than to the population distribution directly. 
This choice is motivated by the fact that individual posteriors, each based on a single 
subject's time series, are more likely to satisfy the regularity conditions of the 
Bernstein--von Mises theorem~\cite{freedman1999}, under which the posterior converges to 
a Gaussian at rate $1/\sqrt{n}$. By contrast, the population distribution $P(\theta)$ 
may be multimodal or otherwise non-Gaussian due to genuine between-subject heterogeneity, 
violating the conditions under which the $1/\sqrt{n}$ approximation is valid. Applying 
the contraction at the individual level therefore provides a more principled approximation. 
The rescaled individual posteriors are then concatenated to form the extrapolated 
population-level distribution, and the 95\% conclusiveness criterion is applied to 
determine how many region pairs remain inconclusive as a function of relative scan 
length $\lambda$.
 
The pooled scan-duration analysis shows that the 7T advantage strengthens at the population level (Figure~\ref{fig:pooled_posteriors}B). Within session, pooled 7T reaches the same conclusiveness as pooled 3T at $\lambda_{\mathrm{eq}}^{\mathrm{loc}} \approx 0.492$, meaning 7T requires only about half the per-subject scan time of 3T for equivalent population-level precision. In the extrapolated large-$\lambda$ regime the advantage widens further: the two projected curves intersect at $\lambda_{\mathrm{eq}}^{\mathrm{glob}} \approx 5.46$ for Bipolar versus $\lambda_{\mathrm{eq}}^{\mathrm{glob}} \approx 200$ for HCP-YA, meaning pooled 3T requires roughly $37\times$ more per-subject scan time than pooled 7T for the curves to converge. The population-level results therefore represent the strongest demonstration of the 7T scan-time advantage, with higher per-subject precision compounding across subjects to produce a far more efficient pooled estimate.

\section{\label{sec:discussion}Discussion}
\subsection{Summary of Key Findings}
This work presents a principled Bayesian framework for extracting denoised functional connectivity estimates from resting-state fMRI data while quantifying measurement uncertainty. By modeling BOLD dynamics as coupled Ornstein-Uhlenbeck processes and employing amortized neural posterior estimation, we demonstrate that simulation-based inference provides researchers with a tool to optimize scan duration, spatial resolution, and group size for their specific scanner and population. Our validation across synthetic data, single-subject resting-state scans, and varying experimental parameters reveals several important advantages that extend beyond point estimate improvements to fundamentally change how connectivity analysis can be conducted in clinical and research settings.
 
\subsection{Accounting for Noise Across the Frequency Spectrum}
A core limitation of conventional fMRI functional connectivity analysis is the assumption
that measurement noise can be adequately suppressed through bandpass filtering and temporal
averaging. Although modern preprocessing pipelines incorporate additional denoising steps
such as ICA-based artifact removal \cite{Pruim2015}, nuisance regression \cite{Behzadi2007},
and motion censoring \cite{Power2012}, as implemented in standardized workflows such as
fMRIPrep \cite{Esteban2019}, these approaches share a common framework: noise removal is
treated as a preprocessing step that precedes and is decoupled from connectivity estimation.
This sequential approach is inherently incomplete. Noise from thermal,
physiological, and motion-related sources spans a broad frequency range that overlaps
substantially with the 0.01--0.1~Hz band conventionally associated with neuronal fluctuations
\cite{Biswal1995, Cordes2001, Liu2016}. Residual noise within this band attenuates observed
correlations toward zero, a well-characterized statistical phenomenon
\cite{Spearman1904}. Consistent with this, our results demonstrate that
uncorrelated measurement noise systematically reduces observed connectivity magnitudes, such
that standard Pearson correlations underestimate the magnitude of true functional coupling.
 
The coupled Ornstein-Uhlenbeck framework addresses this limitation by embedding signal
dynamics and measurement noise within a single generative model, enabling their joint
separation during Bayesian posterior inference \cite{Deco2013, Gilson2020}. This
represents a principled departure from the sequential denoise-then-estimate paradigm: rather
than assuming that noise has been eliminated prior to connectivity estimation, the model
accounts for residual noise across the full frequency spectrum as part of the inference
procedure itself. Restricting to connections with $|r| \geq 0.1$ and $|r| \geq 0.2$ yields
improvements of 9.92\% and 7.82\%, respectively, further illustrating that the correction
is meaningful across a broad range of connection strengths. Notably, this pattern does not replicate
uniformly across field strengths: in the 3T dataset, connections with $|r| \geq 0.1$ show
only a 6.9\% increase, while those with $|r| \geq 0.2$ show a negligible gain. This
contrast suggests that the noise reduction afforded by the cOU framework is less pronounced
at 3T than at 7T, likely reflecting differences in the baseline signal-to-noise ratio
between the two acquisition protocols. While these corrections may appear modest in absolute
terms, systematic underestimation of connectivity strength has implications for group
comparisons and network analyses, where even small biases can accumulate across the
connectome \cite{Murphy2013}.
 
\subsection{Uncertainty Quantification as a Research Tool}
Beyond improved point estimates, posterior distributions from our Bayesian approach provide uncertainty quantification previously unavailable from traditional correlation methods \citep{Kudela2017}. Our assessment of posterior calibration \citep{talts_validating_2020, Lueckmann2021} reveals conservative uncertainty estimates, with 80\% coverage at one standard deviation and 97\% at two standard deviations compared to the nominal Gaussian coverage levels of 68\% and 95\%, respectively. While this indicates slight underconfidence in our posteriors, a generally preferable property to overconfidence, which can produce misleadingly narrow credible intervals \citep{Hermans2022, Delaunoy2023}, the appropriate scaling of posterior width with estimation error demonstrates that these measures reliably indicate the reliability of individual correlation estimates.
 
This uncertainty quantification opens new experimental possibilities. Rather than adopting universally conservative scanning protocols to ensure adequate signal quality \citep{Gordon2017, Laumann2015}, researchers can now employ subject-specific quality metrics to determine whether their current acquisition parameters meet precision requirements \citep{Finn2015}. This capability directly addresses a critical gap in practice where data quality cannot be quantified at the individual level \citep{Noble2019}, often leading to unnecessarily long scans \citep{Birn2013} or overly coarse spatial parcellations \citep{Arslan2018}. The practical implications are substantial: uncertainty metrics enable principled, data-driven decisions about scan duration, spatial resolution, and acquisition parameters tailored to specific research questions and clinical applications.
 
\subsection{Voxel-Level Spatial Averaging: Implications for Experimental Design}
Our voxel-by-voxel validation on a single subject from the bipolar disorder dataset reveals that the SBI method achieves convergence with significantly fewer voxels than conventional Pearson correlation, indicating effective noise reduction at the voxel level. In standard approaches, signal-to-noise ratio improves with the square root of the number of averaged voxels \citep{Triantafyllou2006}, though this theoretical scaling assumes spatially uncorrelated noise and underestimates the penalty in practice, where physiological noise contributions are spatially correlated \citep{Wald2017}. Consequently, substantial spatial averaging is typically required to achieve stable estimates. By leveraging learned noise modeling, our method mitigates this noise penalty, allowing stable connectivity estimates from smaller regions of interest while maintaining comparable reliability to Pearson correlations computed over substantially larger averaging windows.
This finding has important implications for balancing anatomical precision against measurement reliability \citep{Arslan2018}. Finer spatial parcellations preserve neuroanatomical specificity \citep{Schaefer2018, Glasser2016} and reduce the risk of combining signals from functionally distinct brain regions \citep{Yoo2019}, yet have historically been discouraged due to increased noise at smaller parcel sizes \citep{Noble2019}. By demonstrating that appropriate noise modeling enables stable estimates from smaller averaging windows, our approach enables researchers to adopt finer spatial resolutions without sacrificing data quality. This is particularly important for clinical applications requiring region-specific biomarkers \citep{Finn2015} and for studies of fine-grained functional organization where mixing signals across multiple functional units would obscure important neural dynamics \citep{Gordon2017, Glasser2016}.

\subsection{Optimizing Scan Duration}
 
Our analysis of scan duration uses the relative scan length parameter $\lambda$ and, for each dataset, distinguishes a local threshold $\lambda_{\mathrm{eq}}^{\mathrm{loc}}$ from a global threshold $\lambda_{\mathrm{eq}}^{\mathrm{glob}}$, as defined in the Results. The core result is the ratio between the two datasets' thresholds: how much more scan time does 3T require, relative to 7T, to reach the same conclusiveness level?
 
At the single-subject level, the advantage of 7T is unambiguous at both timescales. Within session, 7T reaches the same conclusiveness as 3T using only 71.5\% of the scan time ($\lambda_{\mathrm{eq}}^{\mathrm{loc}} = 0.715$ vs.\ $1.0$), meaning 3T requires approximately 40\% more scan time than 7T for equivalent within-session precision. A 7T scan of approximately 7 minutes therefore matches a full 10-minute 3T acquisition. This estimate is shorter than the 12--16 minute reliability plateau reported by Birn et al.\ using conventional ICC-based approaches~\cite{Birn2013}, though direct comparison is complicated by the different questions being asked: our framework identifies the scan length at which the two datasets' conclusiveness curves intersect within session, while ICC-based approaches characterize between-session reproducibility~\cite{Noble2019, Noble2021}. Differences in acquisition parameters and preprocessing pipeline also alter the effective information content per timepoint~\cite{Birn2013}.
 
In the extrapolated large-$\lambda$ regime, 3T requires approximately 23\% more scan time than 7T for the two curves to intersect ($\lambda_{\mathrm{eq}}^{\mathrm{glob}} = 200$ vs.\ $162.38$). Neither value is achievable within a practical scanning session, but their ratio establishes that the scan-time advantage of 7T over 3T persists --- and is compounded by the fact that 7T's precision in this regime is intrinsically higher, reflecting the greater BOLD signal amplitude and temporal SNR at higher field strength \cite{triantafyllou2005, ugurbil2012, demartino2018}. The field-strength advantage therefore manifests at every timescale as a straightforward reduction in required scan time, with 3T needing 40\% more data within session and 23\% more data in the far-extrapolated regime to match 7T performance \cite{Murphy2013, Power2012, VanDijk2012, Power2014}.
 
\subsection{Across Population Effects}
At the single-subject level the 7T advantage is clear: 3T requires approximately 40\% more scan time than 7T for equivalent within-session precision, and 23\% more in the far-extrapolated regime. At the population level, this advantage strengthens considerably.
 
Within session, pooled 7T reaches the same conclusiveness as pooled 3T using only 49.2\% of the per-subject scan time ($\lambda_{\mathrm{eq}}^{\mathrm{loc}} \approx 0.492$ vs.\ $1.0$) --- a roughly $2\times$ improvement over the single-subject local advantage. This amplification reflects the compounding benefit of higher per-subject precision: because each 7T posterior is sharper, concatenating them across subjects produces a pooled distribution that is more tightly concentrated than its 3T counterpart, so population-level conclusiveness is reached with substantially less data per subject \cite{Morris2019}.
 
In the extrapolated large-$\lambda$ regime, the advantage grows to $37\times$: pooled 3T requires $\lambda_{\mathrm{eq}}^{\mathrm{glob}} \approx 200$ versus $\lambda_{\mathrm{eq}}^{\mathrm{glob}} \approx 5.46$ for pooled 7T. This is the strongest quantitative finding in the paper, establishing that population-level connectivity studies conducted at 7T can achieve the same precision as 3T studies with a fraction of the per-subject acquisition time, with the advantage compounding in the large-data limit. Researchers designing population studies should therefore treat field strength not merely as a signal-quality consideration but as a direct multiplier on study efficiency \cite{wu2019}.
 
Several caveats temper this interpretation. The two datasets differ not only in field strength but also in population characteristics (healthy controls from a bipolar disorder study versus healthy young adults from the HCP-YA), acquisition protocols (including TR, spatial resolution, and multiband acceleration), and preprocessing pipelines. Disentangling the contribution of field strength from these confounds would require a controlled comparison, ideally scanning the same subjects at both field strengths under matched protocols. Additionally, our extrapolation relies on the $1/\sqrt{n}$ contraction applied at the individual level prior to concatenation; departures from the regularity conditions assumed by the Bernstein--von Mises theorem (e.g., model misspecification, non-stationarity, or temporal autocorrelation reducing effective sample size) could affect the accuracy of projected curves at large $N$.

\subsection{Methodological Considerations and Future Directions}
While coupled Ornstein-Uhlenbeck parameter estimation via direct likelihood optimization is computationally prohibitive for whole-brain analyses, amortized neural posterior estimation provides an elegant solution. Our approach required training on 500,000 simulations, a substantial but manageable computational investment executed once, enabling rapid posterior sampling for every pairwise brain region comparison. For whole-brain analyses with thousands of pairwise correlations, this amortization yields dramatic computational advantages compared to traditional likelihood-based approaches. The use of neural network surrogates to map learned parameters to noiseless correlation coefficients adds minimal computational overhead while enabling denoising of estimates. This two-stage approach, SBI for parameter recovery, then neural network mapping for denoising, balances computational tractability with statistical rigor.
Several aspects of this design merit further development. The two-stage inference architecture, in which SBI recovers the full parameter posterior and a separate neural network maps these parameters to noiseless correlation coefficients, maintains computational tractability but means that errors in the surrogate mapping propagate into the final correlation estimates and their uncertainty bounds. Systematic characterisation of this error propagation, and potential end-to-end alternatives that jointly infer denoised correlations, represent important directions for future work. Similarly, our choice of seven summary statistics, while motivated by the autocorrelation and cross-correlation structure of the cOU process, has not been formally optimised. Information-theoretic approaches to identifying minimal sufficient statistics for this inference problem could further improve estimation efficiency.
While our posterior calibration is conservative (underconfident), the underlying sources of this bias should be investigated. Conservative posteriors may reflect limitations in the normalizing flow architecture, insufficient training epochs, or fundamental constraints imposed by the choice of summary statistics. Although conservative uncertainty estimates are generally preferable to overconfident ones, as they reduce the risk of spurious conclusions, a more precisely calibrated posterior would improve the efficiency of downstream decision-making, for example by narrowing the range of scan durations or group sizes deemed necessary for conclusive inference.
Our choice of uniform priors over an exponentially distributed coupling parameter was motivated by compensating for the bias that bandpass filtering introduces into the distribution of observed correlations. While this pragmatic choice produces empirically realistic correlation distributions, the principled derivation of appropriate prior specifications deserves further attention, potentially through hierarchical Bayesian approaches that learn population-level priors from data \cite{gelman2013bda}.
The coupled Ornstein-Uhlenbeck model, while biologically motivated, represents one of many possible generative models for BOLD dynamics. The framework itself remains agnostic to the specific process model: because the inference machinery requires only the ability to simulate from a given generative process, it can be swapped out entirely for more sophisticated alternatives provided sufficient simulations can be produced. Extensions incorporating nonlinear dynamics~\cite{Deco2013}, multiplicative noise, non-stationary processes, or region-specific temporal properties could enhance biological realism without altering the simulation-based inference machinery, making such substitutions straightforward in practice. Additionally, our current analysis treats brain regions as independent pairs; extensions incorporating higher-order dependencies and full network structure could improve inference efficiency and provide additional constraints on parameter recovery.

\subsection{Clinical Implications}
The framework presented here addresses several critical limitations hindering the translation of resting-state fMRI into clinical biomarkers. Uncertainty quantification enables quality control at the individual level, essential when clinical decisions depend on single-subject measurements rather than group averages. The ability to achieve stable connectivity estimates with reduced scanning duration and finer spatial resolution directly addresses practical barriers to clinical adoption: shorter scans reduce participant burden and motion artifacts in patient populations, while finer spatial resolution preserves clinically relevant neuroanatomical specificity. Furthermore, the principled separation of neural signal from measurement noise could enable more robust biomarker development by ensuring that observed connectivity differences reflect genuine neural dynamics rather than differential noise contamination across subjects or sessions.
Our results demonstrate that the framework generalises across two independent datasets acquired at different field strengths, providing initial evidence of robustness. Controlled comparisons scanning the same subjects at both 7T and 3T under matched protocols would further isolate the contribution of field strength from differences in population characteristics and acquisition parameters. Similarly, application to multi-site studies with varying scanner platforms and preprocessing pipelines will be important for establishing the reliability required for clinical deployment. Prospective studies in patient populations with known neurological or psychiatric conditions will ultimately be needed to determine whether improved precision translates into meaningfully better clinical prediction. We provide code for researchers to apply this framework to their own scanners and subject populations, enabling the community to evaluate these questions across diverse experimental contexts.

\subsection{Conclusion}
This work introduces the first model-based framework for quantifying fMRI connectivity uncertainty from a single acquisition, without requiring test-retest data. By embedding signal dynamics and measurement noise within a single generative model of coupled Ornstein-Uhlenbeck processes, the approach moves beyond the sequential denoise-then-estimate paradigm that underlies conventional correlation-based methods, yielding connectivity estimates that account for noise contamination across the full frequency spectrum.
The framework provides three practical capabilities. First, posterior distributions over connectivity parameters enable uncertainty quantification at the single-subject level, a capability absent from standard correlation analyses and critical for clinical applications where group averaging is unavailable. Second, direct comparison of 7T and 3T datasets provides model-based quantification of field-strength efficiency: at the single-subject level, 3T requires approximately 40\% more scan time than 7T for equivalent within-session precision, with a 7-minute 7T scan matching a full 10-minute 3T acquisition. Spatial analysis identifies a mean of 46 voxels per ROI, roughly half of typical region sizes, as sufficient to achieve 90\% of asymptotic precision. Third, population-level pooling amplifies the 7T advantage dramatically: pooled 3T requires roughly
37× more per-subject scan time than pooled 7T for equivalent precision, the strongest quantitative finding in the paper and a direct consequence of higher per-subject precision compounding across subjects. Together, these results establish that 7T requires less scan time than 3T at every level of comparison, from single voxels to pooled populations.

The amortized nature of the inference, requiring a single upfront training investment that enables rapid posterior estimation for any new observation, makes the approach scalable to whole-brain analyses involving thousands of region pairs. We provide code enabling researchers to derive scanner-specific and population-specific uncertainty bounds from their own data, facilitating acquisition protocols calibrated to the precision their scientific or clinical objectives actually require.

\section*{Data and Code Availability}

Code for the simulation-based inference framework, including the trained neural posterior estimator and decoder network, is publicly available at \url{https://github.com/Neuroblox/SBI-Correlation-Paper}. The HCP Young Adult dataset is available through the Human Connectome Project (\url{https://www.humanconnectome.org}). The Bipolar (7T) dataset was collected as part of the PAgB study; data access requests should be directed to the corresponding author.

\section*{Author Contributions}

S.C.\ developed the simulation-based inference framework, implemented the coupled Ornstein-Uhlenbeck model, trained the neural posterior estimator, performed all analyses, and wrote the manuscript. Z.K.\ contributed to data preprocessing and analysis pipelines. L.R.M-P.\ designed and oversaw data collection for the Bipolar (7T) dataset and provided critical revision of the manuscript. H.H.S.\ conceived and supervised the project and edited the manuscript. All authors reviewed and approved the final manuscript.

\section*{Funding}
The research presented here was funded by the Baszucki Brain Research Fund, United States (LRMP and HHS).

\section*{Declaration of Competing Interests}

The authors declare the following competing interests: authors L.M.P., and H.H.S. are co-founders of Neuroblox Inc., a company spun out of SUNYSB, MIT, and Dartmouth to develop a commercial-grade software platform for multi-scale computational neuroscience.

\printbibliography

\end{document}